\documentclass[letterpaper]{article} 
\usepackage{aaai25}  
\usepackage{times}  
\usepackage{helvet}  
\usepackage{courier}  
\usepackage[hyphens]{url}  
\usepackage{graphicx} 
\usepackage{natbib}  
\usepackage{caption} 
\usepackage{algorithm}
\usepackage{algorithmic}
\usepackage{soul}
\usepackage{newfloat}
\usepackage{listings}
\usepackage{amsfonts,amssymb}
\usepackage{amsmath}
\usepackage{booktabs}
\usepackage[table]{xcolor}
\usepackage{ragged2e}
\usepackage{booktabs,makecell, multirow, tabularx}
\usepackage{arydshln}
\DeclareCaptionStyle{ruled}{labelfont=normalfont,labelsep=colon,strut=off} 
\floatstyle{ruled}
\newfloat{listing}{tb}{lst}{}
\floatname{listing}{Listing}
\title{VA-AR: Learning Velocity-Aware Action Representations with \\ Mixture of Window Attention}
\author{
    Jiangning Wei\textsuperscript{\rm 1}\equalcontrib,
    Lixiong Qin\textsuperscript{\rm 1}\equalcontrib,
    Bo Yu\textsuperscript{\rm 1},
    Tianjian Zou\textsuperscript{\rm 1},
    Chuhan Yan\textsuperscript{\rm 2},
    Dandan Xiao\textsuperscript{\rm 3},
    Yang Yu\textsuperscript{\rm 4},
    Lan Yang\textsuperscript{\rm 1},
    Ke Li\textsuperscript{\rm 1},
    Jun Liu\textsuperscript{\rm 1}\thanks{*Corresponding author}
}
\affiliations {
    \textsuperscript{\rm 1}Beijing University of Posts and Telecommunications\\
    \textsuperscript{\rm 2}Macau University of Science and Technology\\
    \textsuperscript{\rm 3}China Institute of Sport Science\\
    \textsuperscript{\rm 4}Beijing Sport University\\
    \{trinity, lxqin, liujun\}@bupt.edu.cn
}

\begin{document}
\maketitle

\begin{abstract}
Action recognition is a crucial task in artificial intelligence, with significant implications across various domains. We initially perform a comprehensive analysis of seven prominent action recognition methods across five widely-used datasets. This analysis reveals a critical, yet previously overlooked, observation: as the velocity of actions increases, the performance of these methods variably declines, undermining their robustness. This decline in performance poses significant challenges for their application in real-world scenarios. Building on these findings, we introduce the Velocity-Aware Action Recognition (VA-AR) framework to obtain robust action representations across different velocities.
Our principal insight is that rapid actions (\textit{e.g.}, the giant circle backward in uneven bars or a smash in badminton) occur within short time intervals, necessitating smaller temporal attention windows to accurately capture intricate changes. Conversely, slower actions (\textit{e.g.}, drinking water or wiping face) require larger windows to effectively encompass the broader context. VA-AR employs a Mixture of Window Attention (MoWA) strategy, dynamically adjusting its attention window size based on the action's velocity. This adjustment enables VA-AR to obtain a velocity-aware representation, thereby enhancing the accuracy of action recognition. Extensive experiments confirm that VA-AR achieves state-of-the-art performance on the same five datasets, demonstrating VA-AR's effectiveness across a broad spectrum of action recognition scenarios.
\end{abstract}

\begin{links}
    \link{Code}{github.com/TrinityNeo99/VA-AR_official}
\end{links}

\section{Introduction}
Action recognition serves as a pivotal subfield within the domain of computer vision, encompassing a wide array of practical applications and attracting significant research attention due to its unique challenges. The task demands that models are capable of processing intricate spatial information while efficiently deciphering dynamic temporal data~\cite{herath2017going,zhang2019comprehensive,liu2023dual}. 
With the rapid progress of deep learning technologies, substantial strides have been made in the field of action recognition, particularly owing to the efficiency of skeleton data representation~\cite{ahmad2021graph,xin2023transformer} and the breakthroughs in human pose estimation techniques~\cite{openpose,hrnet,yolov8,mediapipe}.
Skeleton-based action recognition has demonstrated enormous potential for application in various domains, including video surveillance, VR/AR gaming, and sports analytics, significantly enhancing daily life, entertainment experiences, and workflows.

\begin{figure}[tb] 
\centering 
\includegraphics[width=\columnwidth]{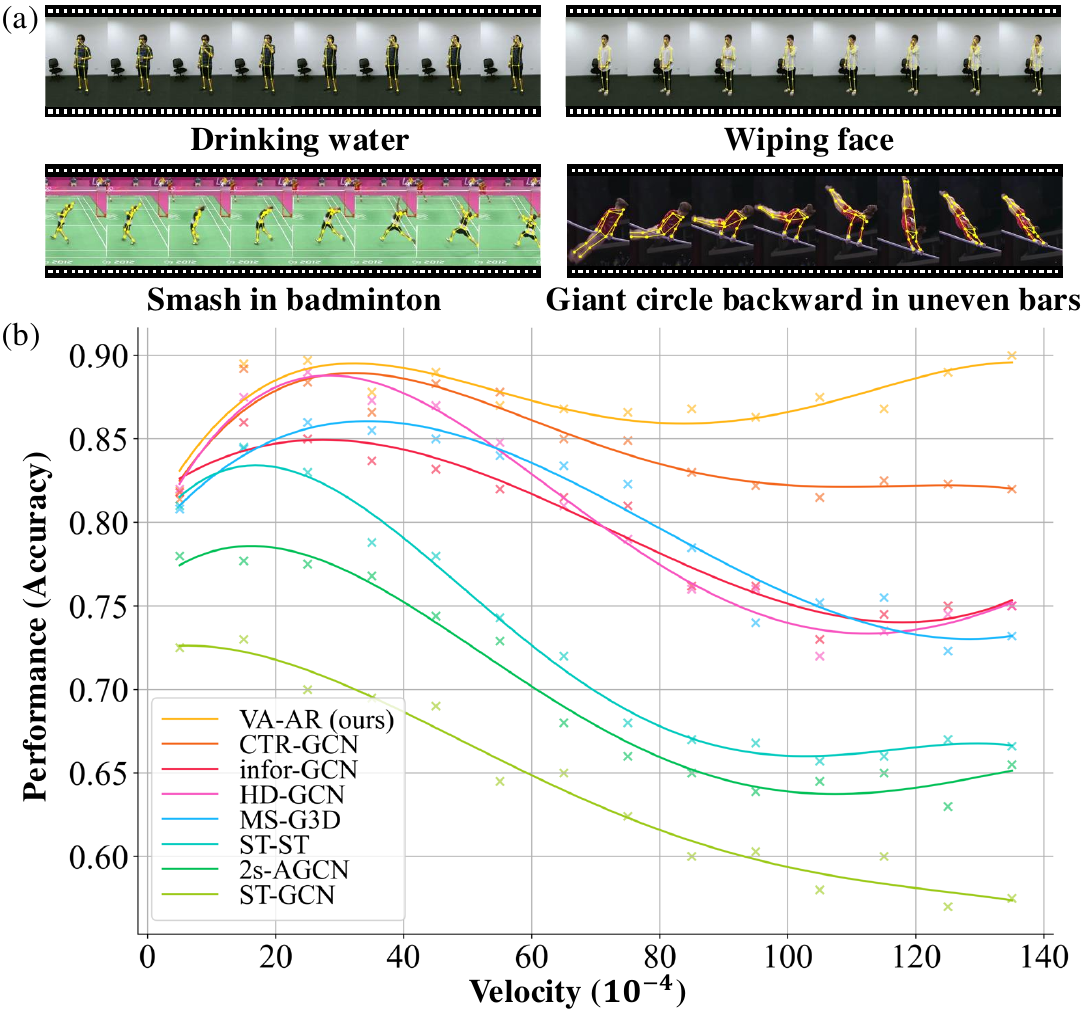} 
\caption{(a) Example videos demonstrating actions at diverse velocities. (b) Performance-velocity curves for seven prominent action recognition methods and our proposed VA-AR, evaluated across five widely-used datasets.}
\label{Fig:conception} 
\end{figure}

Following the captivating events of the 2024 Paris Olympics, our interest in the versatility of existing skeleton-based action recognition algorithms intensified, particularly in their ability to be applied to both daily action prediction and intense athletic action recognition. 
Given the vast diversity of behavior categories in both daily activities and athletic sports, a category-by-category investigation lacks generality. Therefore, we introduced a novel dimension—\textit{Velocity}—to comprehensively assess the performance of behavior recognition algorithms. For instance, daily activities such as drinking water or wiping face exhibit similar slow velocities, whereas sports like a smash in badminton and the giant circle backward in uneven bars demonstrate faster velocities, as shown in Fig.~\ref{Fig:conception} (a). Specifically, we reclassified the data based on the average velocity of individuals performing the actions, maintaining consistency in the data distribution within each category owing to the similarity of average velocities within each category. Additionally, the finite nature of human speed ensures the universality of this classification method.
After merging multiple datasets, which encompass both daily and athletic behaviors, and statistically analyzing the recognition accuracy of various algorithms based on behavior speed, the results were disheartening. As illustrated in Fig.~\ref{Fig:conception} (b), when behavior speed increases, the performance of existing algorithms noticeably deteriorates. This phenomenon suggests that existing algorithms exhibit evident limitations when faced with actions of different speeds.

In the domain of action recognition, behaviors with slower speeds exhibit relatively minor differences between consecutive frames, whereas those with faster speeds demonstrate more pronounced inter-frame discrepancies. 
This characteristic is encapsulated within the temporal module, where slower behaviors demand a broader temporal receptive field to aggregate behavioral information from a wider spectrum of frames, whereas faster behaviors require a more focused temporal receptive field to concentrate on local temporal nuances.

In the context of skeleton-based action recognition tasks, the design of the temporal module has primarily been reliant on Temporal Convolutional Networks (TCNs) \cite{lea2017temporal} or Transformer \cite{vaswani2017attention} architectures, which have yielded significant research accomplishments. TCNs and their multi-scale variants \cite{cui2016multi,li2020ms} learn temporal relationships between frames through convolutional operations. However, due to the inherent weight-sharing mechanism of convolutional networks, frames with the same relative position are assigned uniform weights, resulting in a fixed granularity of temporal modeling that fails to adapt to the specific requirements of individual behavior samples. Consequently, these methods exhibit robust performance within a specific speed range but degrade significantly in performance beyond that range.
Conversely, Transformers, through their multi-head self-attention mechanism, are capable of dynamically adjusting the granularity of attention to token-level, effectively capturing global long-term temporal dependencies. However, this mechanism also entails an increase in computational complexity, and the model must learn an extensive amount of information—requiring it to dynamically focus on local details while also attending to global semantic information. This ultimately leads to a less pronounced performance of Transformers in skeleton-based action recognition tasks \cite{zhang2022zoom,plizzari2021spatial}, resulting in a relatively low adoption rate in this domain.

We are focused on developing an algorithm that is insensitive to velocity variations and can dynamically attend to critical details. Therefore, we have established Transformer as the foundational architecture for our temporal module. Inspired by the success of Swin Transformer \cite{liu2021swin} in visual tasks, which reduces computational burden significantly and clearly guides the model to focus on key local features through a local window self-attention mechanism, we have applied this strategy to skeleton-based action recognition tasks. By unfolding the temporal dimension of the skeleton sequence’s spatiotemporal features and uniformly or shifted dividing them into different windows, we perform self-attention only within the windows. This strategy, through Local Window Attention (LWA), explicitly constrains the Transformer to focus on local temporal information, while Shifted Window attention (SWA) allows the model to attend to global semantic information. After introducing the window division strategy, we are able to draw on successful experiences from convolutional neural networks, such as the application of multi-scale TCN in action recognition \cite{liu2020disentangling, qin2022fusing}. We propose multi-scale window attention, which enables the model to focus on temporal information of different lengths, thereby enhancing its temporal modeling capability and compatibility with different action velocities. To further enhance the model’s adaptive capability in focusing on the granularity of the temporal dimension, we decide to employ expert modules to provide decision information. Through the Mixture of Experts (MoE) model, we provide token-level attention weights and use a routing network composed of a linear mapping layer to decide the activation state of the expert models. To reduce computational load, we reuse LWA and SWA, allocating a single window scale to an expert, so that our expert models only require an additional computational burden of one linear layer. This design not only improves the model’s efficiency but also maintains its flexibility and accuracy in handling actions at different velocities.

Our main contributions are as follows:

\begin{itemize}
    \item We conduct a study on the impact of action velocity and observed that current skeleton-based methods exhibit significant performance degradation with varying action velocities, particularly for high-speed actions, thus failing to achieve robust action recognition.
    \item We thus introduce the Mixture of Window Attention (MoWA), which can be integrated into the Velocity-Aware Transformer to obtain robust action representations across different velocities. Building on this, we propose the Velocity-Aware Action Recognition (VA-AR) framework to facilitate end-to-end action recognition.
    \item Extensive experiments validate the effectiveness of our VA-AR, which achieves state-of-the-art performance on five widely used datasets, spanning various action velocities.
\end{itemize}

\begin{figure*}[htbp]
    \centering
    \includegraphics[width=0.95\textwidth]{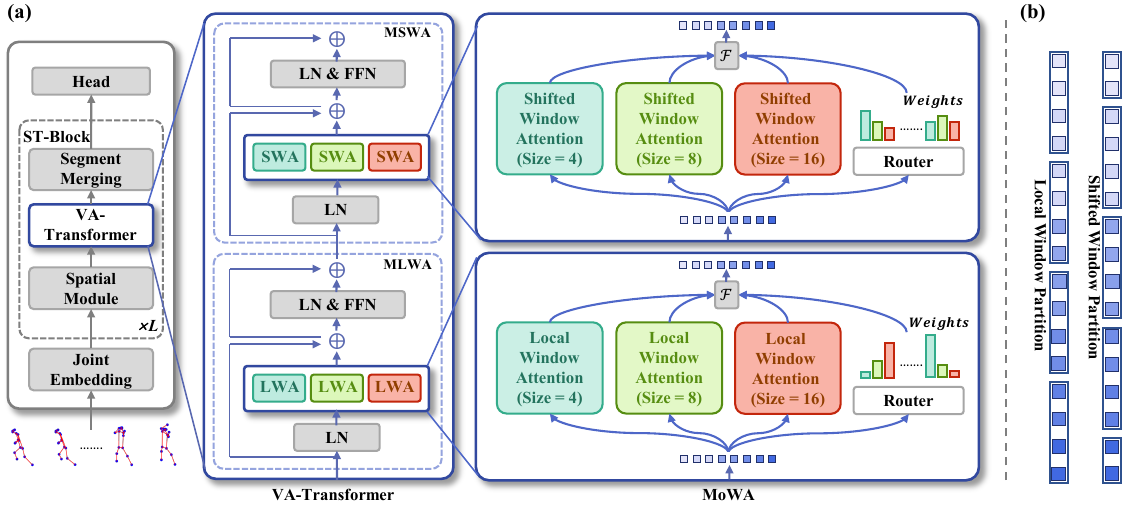} 
    \caption{(a) The Architecture of VA-AR. The Velocity-Aware Transformer incorporates the Mixture of Window Attention (MoWA), implemented as Multi-scale Local Window Attention (MLWA) and Multi-scale Shifted Window Attention (MSWA). Within this framework, MoWA is embedded to dynamically adjust the attention window weights across various scales, facilitating adaptation to changes in action velocity. (b) Different window partitioning strategies, with a window size of 4.}
    \label{fig: architecture}
\end{figure*}

\section{Related Works}
\subsection{Velocity in Skeleton-based Action Recognition}
Velocity, as a critical attribute of human activity, was first introduced as a novel modality for model ensemble in the study~\cite{song2020stronger}. The velocity feature is derived by subtracting adjacent frames prior to training. 
The effectiveness of motion features utilized for model ensemble has been validated by several studies~\cite{liu2020disentangling, chi2022infogcn, chen2021channel, zhou2024blockgcn, lee2023hierarchically, qin2022fusing}. However, previous studies have treated motion features as auxiliary information to enhance model performance, employing the same model architecture for both motion and original joint features. This straightforward approach lacks a meticulous design tailored for velocity, thereby limiting the representation learning. To fully exploit the benefits of velocity and address these limitations, we propose VA-AR, which adaptively adjusts to the temporal granularity, ensuring effective velocity-aware representation learning.

\subsection{Temporal Relation Modeling of Skeleton-based Action Recognition}
In the field of skeleton-based action recognition, existing methods extract spatiotemporal features from input skeleton sequences and project them onto category scores. 
The learning of these representations can be categorized into spatial and temporal components.
Despite the advancements for spatial modeling, 
the learning approaches in the temporal domain remain relatively simplistic, primarily focusing on Temporal Convolutional Networks (TCN) and Transformer models. Earlier methods, such as ASGCN~\cite{yu2021gcn}, 2s-AGCN~\cite{shi2019two}, and ms-AGCN~\cite{shi2020skeleton}, utilize TCN for temporal representation by performing convolution operations along the temporal dimension of each joint. Inspired by the success of multi-scale features in CNNs~\cite{szegedy2015going, lin2017feature, girshick2015fast}, several studies~\cite{liu2020disentangling, zhou2024blockgcn, lee2023hierarchically, qin2022fusing} have expanded TCN into multi-scale TCN to fuse features of different granularities. Due to the static nature of these convolution kernels, TCN cannot adaptively focus on critical frames within the skeleton sequence, thereby limiting model performance. To address this, some works~\cite{zhang2022zoom, plizzari2021spatial} have adopted Transformers to model joint relationships across different frames. However, the global nature of the temporal Transformers in these studies disregards the locality of adjacent frames and introduces substantial computational complexity. To mitigate these challenges and harness the strengths of both TCN and Transformer, we propose the local window attention and shifted window attention modules to adaptively focus on significant skeleton frames.

\subsection{Mixture of Expert} 
The Mixture of Experts (MoE) model was first introduced in the 1990s~\cite{jacobs1991adaptive}. ~\citeauthor{shazeer2017outrageously} later integrated MoE into modern deep learning frameworks, demonstrating its effectiveness in Language Modeling and Machine Translation. 
Recently, MoE has gained popularity in Large Language Models (LLMs) such as LLaMA-MoE~\cite{zhu2024llama}, Mixtral-8x7B~\cite{jiang2024mixtral}, Qwen2~\cite{yang2024qwen2}, and Zephyr Orpo 141B~\cite{hong2024orpo}. Additionally, the Mixture of Attention (MoA) was proposed by~\citeauthor{zhang2022mixture}, which combines multi-head self-attention with the MoE mechanism.
Specifically, multiple attentions are calculated initially and subsequently mixed by a token-wise routing network. 
The initial motivation behind MoE and MoA was to increase the model's parameter size without significantly boosting computational complexity.
Beyond NLP, MoE has also emerged as a promising technology in image segmentation~\cite{liu2020att}, image generation~\cite{ostashev2024moa}, and sequential recommendation~\cite{cho2020meantime}. However, in this paper, our motivation for employing MoA is to align skeleton frames with the most suitable window attention, thereby enhancing the model’s adaptive capability in focusing on the granularity of the temporal dimension.

\section{Methodology}
\subsection{Overall Architecture}
The VR-AR model is designed to learn robust features that are insensitive to variations in speed, adaptively focusing on the key local and global information of actions at different velocities, thereby enhancing the recognition accuracy for diverse types of behaviors. 
The architecture of the VR-AR model is illustrated in Fig.~\ref{fig: architecture} (a). The model takes a sequence of skeleton coordinates $C_{N, T} \in \mathbb{R}^{N \times T \times C}$, with $N$ joints and $T$ frames, as input data and transforms them into a latent embedding through a joint embedding module.
Subsequently, the latent embedding is processed by a series of spatiotemporal blocks (ST-Blocks) to extract velocity-aware spatiotemporal features. Ultimately, these features are effectively mapped to a semantic space by the classification head, achieving the categorization of actions. It is noteworthy that the design of the ST-Block comprises three main components: a spatial module, a Velocity-Aware Transformer, and a segment merging module. The spatial module is responsible for extracting spatial features from the latent embedding, while the Velocity-Aware Transformer ensures the model’s stable feature extraction capability across different velocities of actions. The segment merging module is used to integrate features from different segments.

\subsection{Velocity-Aware Transformer}
The Transformer architecture has demonstrated exceptional capability in processing temporal information for action recognition; however, it often focuses on coarse-grained parsing of long-range dependencies, neglecting the attention to local temporal details. To address this issue, we ingeniously extend the concepts of local windows attention and shifted window attention strategies, originally designed for spatial features in the Swin Transformer \cite{liu2021swin}, to temporal features. This extension enables the model to concentrate more effectively on the parsing of local temporal nuances while maintaining the ability to represent long-range dependencies.

Although the attention strategies significantly enhance the model's capacity to capture local temporal nuances, they do not adequately address the requirement for diverse local temporal receptive fields in recognizing actions at varying velocities. To this end, we innovatively propose the \textbf{Multi-scale Local Window Attention (MLWA)} and \textbf{Multi-scale Shifted Window Attention (MSWA)} modules to meet the demands of action recognition at different velocities. Furthermore, we introduce a \textbf{Mixture of Window Attention (MoWA)} model, which can adaptively focus on different temporal receptive fields, thereby ensuring the model's compatibility with a wide range of action recognition tasks across various velocities.

\paragraph{Multi-scale Local Window Attention}
Within the framework of the self-attention mechanism, we adopt a local window strategy, which computes self-attention solely within the confines of a window. This approach effectively reduces computational complexity and guides the model to focus on analyzing local temporal information. Specifically, the spatial features $f \in \mathbb{R}^{N \times T \times d}$ of the skeleton sequence are evenly partitioned along the temporal dimension into $f$ feature segments $\{f_0, f_1,..., f_{\lfloor T/s \rfloor-1}\}, f_i \in \mathbb{R}^{N \times s \times d}$, where $s$ is the size of local windows, and $d$ is the dimension of the spatial feature. This partitioning process begins from the initial frame of the input sequence, ensuring that there is no overlap between the segments, as depicted in Fig.~\ref{fig: architecture} (b). 
The local window feature sequences utilize a multi-head self-attention mechanism to facilitate an in-depth exploration and targeted focus on the features within the segments, as illustrated by the following formula:
\begin{equation}
\small
\begin{aligned}
{\rm LWA}(f,s) &= {\rm Concat}({\rm SA}(f_i)), i \in {0,1,...,\lfloor T/s \rfloor-1},\\
\end{aligned}
\label{eq:LWA}
\end{equation}
where $\rm SA $ and ${\rm Concat}$ denote the operations of Multi-Head Self-Attention and Concatenate.

However, a single window size cannot meet the diverse requirements for local temporal granularity in recognizing actions at different velocities. We implement a local window strategy with multiple window sizes, as illustrated in Fig.~\ref{fig: architecture} (a), allowing the model to adapt flexibly to different window size requirements and effectively recognize actions at various velocities. The formula of \textbf{MLWA} is as following:
\begin{equation}
\small
\begin{aligned}
{\rm MLWA}(f) &= {\rm Concat}({\rm LWA}(f,s_k)),k \in {0,1,...,K-1},\\
\end{aligned}
\label{eq:MLWA}
\end{equation}
where $K$ is the number of window sizes, and $s_k$ is the $k$-th window size of the multi-scale local window attention. 
${\rm Concat}$ denotes the Concatenate.

\begin{table*}[htbp]
  \renewcommand\arraystretch{0.85}
  \tabcolsep=0.1cm
  \scriptsize
  \centering
    \begin{tabular}{lllcccccccc}
    \toprule
    \multirow{3}[4]{*}{\textbf{Method}} & \multicolumn{3}{c}{\textbf{Comments}} & \multicolumn{7}{c}{\textbf{Dataset}} \\
\cdashline{2-11}        & \multirow{2}[2]{*}{Summary} & \multirow{2}[2]{*}{Architecture} & \multirow{2}[2]{*}{Modalities} & \multicolumn{2}{c}{NTU-60} & \multicolumn{2}{c}{NTU-120} & \multirow{2}[2]{*}{P2A} & \multicolumn{1}{c}{\multirow{2}[2]{*}{\makecell{Olympic\\Badminton}}} & \multirow{2}[2]{*}{FineGym} \\
\cdashline{5-8}
          &       &       &       & X-Sub & X-View & X-Sub & X-Set &       &       &  \\
    \midrule
    ST-GCN~\cite{yan2018spatial} & Spatiotemporal Graph Learning & GCN+TCN & J     & 81.5  & 88.3  & 70.7  & 73.2  & 60.1  & 58.9  & 25.0  \\
    AS-GCN~\cite{li2019actional} & Extra Joint Action Links  & GCN+TCN & J     & 86.8  & 94.2  & 78.3  & 79.8  & 65.6  & 63.7  & 80.2  \\
    2S-AGCN~\cite{shi2019two} & Learable Topology & GCN+TCN & J+B   & 88.5  & 95.1  & 82.5  & 84.2  & 70.1  & 69.9  & 84.8  \\
    MS-G3D~\cite{liu2020disentangling}& Temporal Awarded Spatial Modeling & GCN+MS-TCN & J+B+JM+BM & 91.5  & 96.2  & 86.9  & 88.4  & 72.5  & 72.3  & 90.1  \\
    CTR-GCN~\cite{chen2021channel} & Channel-wise Topology Refinement  & GCN+MS-TCN & J+B+JM+BM & 92.4  & 96.8  & 88.9  & 90.6  & 73.0  & 72.2  & 89.6  \\
    ST-TR~\cite{plizzari2021spatial} & Spatiotemporal Transformer & Trans.+Trans. & J+B & 89.9  & 96.1  & 81.9  & 84.1  & 68.0  & 67.1  & 85.1  \\
    ASE-GCN~\cite{qin2022fusing} & Angular Feature & GCN+MS-TCN & J+B+A & 91.6  & 96.3  & 88.2  & 89.2  & 71.8  & 70.4  & 87.1  \\
    InfoGCN~\cite{chi2022infogcn}& Information Bottleneck Loss & GCN+MS-TCN & J+B+JM+BM & 93.0  & 97.1  & 89.8  & 91.2  & 69.6  & 69.4  & 89.9  \\
    ZoomTrans~\cite{zhang2022zoom}* & Multi-person Joint Modeling & Trans.+MS-TCN & J     & 86.5  & 92.1  & 80.1  & 81.7  & 69.9  & 68.0  & 81.1  \\
    ZoomTrans~\cite{zhang2022zoom} & Multi-person Joint Modeling & Trans.+MS-TCN & J+B+JM+BM & 90.1  & 95.3  & 84.8  & 86.5  & 70.5  & 68.5  & 83.4  \\
    HD-GCN~\cite{lee2023hierarchically} & Hierarchical Joint Decomposition & GCN+MS-TCN & J+B+JM+BM & 93.0  & 97.0  & 85.7  & 87.3  & 71.4  & 72.3  & 90.5  \\
    BlockGCN~\cite{zhou2024blockgcn} & Bone Topology Refinement & GCN+MS-TCN & J+B+JM+BM & \textbf{93.1 } & 97.0  & \textbf{90.3 } & \textbf{91.5 } & 72.1  & 72.2  & 86.6  \\
    DS-GCN~\cite{xie2024dynamic} & Dynamic Semantic GCN & GCN+MS-TCN & J+B+JM+BM & \textbf{93.1} & \textbf{97.5} & 89.2 & 91.1 & 71.0 & 70.7 & 90.3 \\
    \hdashline
    \rowcolor[rgb]{ .851,  .851,  .851} Ours  & Mixture of Multi-Window Attention & Trans.+VA-Trans. & J     & 88.7  & 93.5  & 84.6  & 87.1  & 72.9  & 72.3  & 90.0  \\
    \rowcolor[rgb]{ .851,  .851,  .851} Ours  & Mixture of Multi-Window Attention & GCN+VA-Trans. & J     & 90.1  & 94.6  & 86.2  & 89.1  & 73.5  & 73.2  & 91.2  \\
    \rowcolor[rgb]{ .663,  .816,  .557} Ours  & Mixture of Multi-Window Attention & GCN+VA-Trans. & J+B+JM+BM & \textbf{93.1 } & 97.2  & \textbf{90.3 } & \textbf{91.5 } & \textbf{73.9 } & \textbf{73.4 } & \textbf{92.8 } \\
    \bottomrule
    \end{tabular}%
  \caption{Performance comparisons of action recognition on NTU-60, NTU-120, P2A, Olympic Badminton, and FineGym datasets. Modalities are denoted as Joint (J), Bone (B), Joint Motion (JM), Bone Motion (BM), and Bone Angular (A) respectively. For equitable comparisons, performance metrics are reported for models trained using both the single Joint modality and an ensemble of the four modalities, and * represents our implementation.}
  \label{tab:performance}%
\end{table*}%

\paragraph{Multi-scale Shifted Window Attention}
In adopting the local window attention strategy to enhance the model's ability to capture local temporal nuances, we recognized that this strategy may overlook the modeling of long-range dependencies. To overcome this limitation, we introduce the shifted window attention strategy. In the shifted window attention, the starting frame of the sequence is adjusted to the $\lfloor s/2 \rfloor$-th frame, and non-overlapping partitions are executed, as depicted in Fig.~\ref{fig: architecture} (b). Specifically, the spatiotemporal features $f^{'}$ extracted by MLWA are divided into $\lfloor T/s \rfloor +1$ segments, denoted as ${ f^{'}_0, f^{'}_1, ..., f^{'}_{\lfloor T/s \rfloor}}$. Among these segments, $f^{'}_0$ and $f^{'}_{\lfloor T/s \rfloor} \in \mathbb{R}^{N \times s/2 \times d}$, while the remaining segments $\{ f^{'}_1, ..., f^{'}_{\lfloor T/s \rfloor-1}\}, f^{'}_i \in \mathbb{R}^{N \times s \times d}$. This shifted window partitioning allows information within the sequence to flow freely, thereby facilitating the effective extraction of global temporal information. 

The multi-scale windows size strategy and the multi-head self-attention mechanism within windows are also applied to the shifted windows. The formula expression of the \textbf{MSWA} process is as follows:
\begin{equation}
\small
\begin{aligned}
{\rm MSWA}(f^{'}) &= {\rm Concat}({\rm SWA}(f^{'},s_k)),k \in {0,1,...,K-1},\\
{\rm SWA}(f^{'},s_k) &= {\rm Concat}({\rm SA}({f^{'}}^k_i)),i \in {0,1,...,\lfloor T/s \rfloor}.\\
\end{aligned}
\label{eq:MSWA}
\end{equation}
This integration not only retains the local window attention strategy’s fine-grained attention to local temporal nuances but also effectively captures long-range dependencies, i.e., global temporal information, through the sliding mechanism of the shifted window.

\paragraph{Mixture of Window Attention}
Actions of different speeds have distinct requirements for the temporal receptive field. Although a multi-scale window strategy can accommodate various speeds of action, for specific action samples, the model should be able to adaptively select an appropriate temporal receptive field. To address this issue, we propose a MoE model based on attention mechanisms. This model can dynamically adjust the size of the temporal receptive field based on the characteristics of the action samples.

In our MoWA, $M$ experts are responsible for per window size, thereby forming a module comprising $M \cdot K$ experts. The network structure of the expert model is the same as LWA or SWA. In practical applications, the expert modules were implemented using the LWA and SWA modules from the VA-Transformer. This approach of repurposing existing components not only streamlines the model’s architectural complexity but also enhances its computational efficiency.
To enable the selection of the expert models, a routing network $G$, composed of linear layers, maps the current features $f$ to the probability $G(f) \in \mathbb{R}^{N \times T \times (M\cdot K)}$ of the expert models being selected with:
\begin{equation}
\small
\begin{aligned}
G(f) = {\rm Softmax}(f \cdot W_g),
\end{aligned}
\label{eq:G}
\end{equation}
where $W_g \in \mathbb{R}^{d \times (M \cdot K)}$ is the parameters of routing network $G$. $G(f)$ is capable of dynamically selecting the most suitable expert models to process temporal information at various scales.

To integrate the MoE with the multi-scale window strategy, we employ a weighted sum method, treating the probability distribution $G(f)$ as the weights for features output by $M \cdot K$ experts. This can be expressed by the following formula:
\begin{equation}
\small
\begin{aligned}
f_{\rm{MoWA}} = G(f)*{\rm Concat}({\rm LWA}_m(f,s_k))\\
or\\
f_{\rm{MoWA}} = G(f^{'})*{\rm Concat}({\rm SWA}_m(f^{'},s_k))\\
k \in {0,1,...,K-1}, m \in {0,1,...,M-1}. \\
\end{aligned}
\label{eq:MoE}
\end{equation}

Following Layer Normalization (LN), the feature $f_{\rm{MoWA}}$ is further processed by a Multi-Layer Perceptron (MLP) and is then passed on to subsequent modules for further processing. We selected the classic cross-entropy loss function as the final optimization objective for VA-AR.

\begin{figure*}[ht!] 
\centering 
\includegraphics[width=0.99\textwidth]{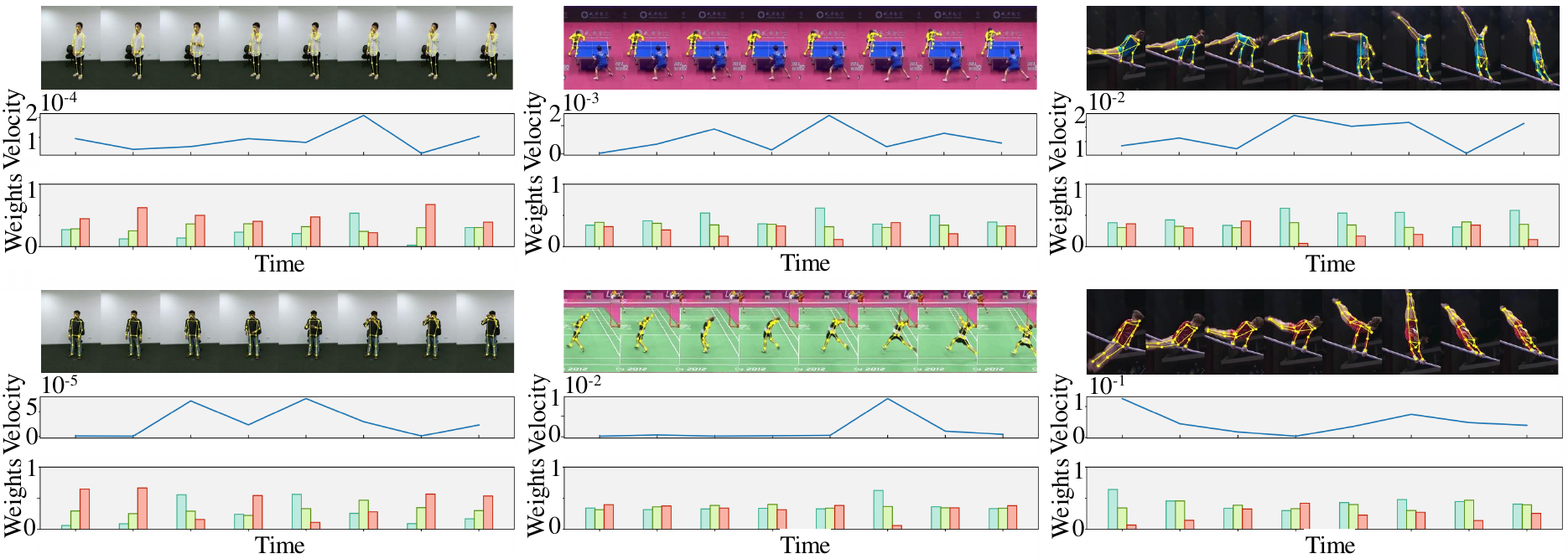} 
\caption{Visualization of velocity-time curves and the corresponding weights generated by our Mixture of Window Attention (MoWA) for each frame across varying velocities. It is noticed that the scales of the velocity of these samples are different. (Right column: slow actions; Middle column: middle-speed actions; Left column: fast actions)}
\label{fig:visualization} 
\end{figure*}

\section{Experiment}

\subsection{Datasets}
We meticulously selected five representative action recognition datasets to comprehensively assess the performance of the proposed method. These datasets comprise NTU RGB+D (NTU-60) \cite{shahroudy2016ntu}, NTU RGB+D (NTU-120) \cite{liu2019ntu}, P2A \cite{bian2022textbf}, Olympic Badminton \cite{ghosh2018towards}, and FineGym \cite{shao2020finegym}. The NTU-60 and NTU-120 datasets are primarily centered on the recognition of actions in daily life, characterized by relatively consistent variations in the speed of action execution. In contrast, the P2A, Olympic Badminton, and FineGym datasets encompass a variety of athletic sports scenarios, where the complexity of action recognition is attributed to the rapid fluctuations in movement velocities, particularly the dynamic capturing of skeleton keypoint positions. Detailed information on data processing and additional experimental analyses can be found in the supplementary material.

\subsection{Implementation Details}
We have selected to use the Graph Convolutional Network (GCN) as the Spatial Module and have configured three ST-Blocks. Additionally, we have employed three different window sizes of 4, 8, and 16. During the experimental process, we employed two NVIDIA 3090 GPUs for training, which encompassed a total of 60 training epochs. For the optimizer, we have chosen SGD with a momentum of 0.9 and a weight decay parameter of 0.0001, with batch size setting to 32. The maximum temporal lengths for the NTU-60, NTU-120, and FineGym datasets were set to 256; whereas, for the P2A and Olympic Badminton datasets, the maximum temporal lengths were set to 128. 
The velocity is defined using a three-frame joints calculation (previous, current, and subsequent frames) for analysis, with detailed explanations provided in the supplementary material.

\subsection{Quantitative Results}
\paragraph{Baseline Comparison} We conducted a comprehensive comparison with 11 methods based on skeleton data for action recognition. The experiments covered five large datasets, with published results copied into the paper, and unpublished results obtained through retraining on these datasets. We conducted rigorous validations on both Joint data and multi-modalities data (e.g., Joint, Bone, Joint Motion, and Bone Motion). 

From Tab.~\ref{tab:performance}, we can make the following observations: i) Our method has demonstrated exceptional performance across all datasets, encompassing both single-modality and multi-modalities data. This not only validates the efficacy of our proposed method but also attests to its broad applicability. ii) On sports activity datasets—P2A, Olympic Badminton, and FineGym, our method achieved significant performance gains over the compared methods, with improvements of $1.23\%, 1.52\%, 2.54\%$, respectively. This enhancement is notably greater than that observed on daily behavior datasets—NTU 60 and NTU 120. This discrepancy may stem from the fact that the model architectures of the compared methods are primarily designed for slow-paced daily behaviors and, consequently, are less capable of focusing on fine-grained temporal information within small scopes when dealing with high-speed sports activities. iii) The addition of modalities resulted in a more pronounced performance improvement on daily behavior datasets, while on sports activity datasets, the increase in performance was not as significant. This could be because sports activity datasets, due to the high-speed movements involved, require smaller temporal windows to capture local temporal nuances. These smaller windows can easily learn motion information calculated from frame differences, hence the additional motion modalities do not offer substantial improvement for these datasets. iv) Although the Transformer architecture has demonstrated its formidable capabilities in various computer vision tasks, its extensive application and outstanding performance in the field of action recognition have not been fully exploited. We aim to bridge this gap by introducing innovative modules---Velocity-Aware Transformer, successfully applying the MoWA to the Transformer architecture, and achieving significant results. 

\paragraph{Velocity Insensitivity}
We conducted a thorough quantitative analysis of the impact of action speed on action recognition. We mixed the test sets of the five datasets and carefully distinguished them based on action speed. The specific method for calculating speed is detailed in the supplementary materials, and the trend of accuracy in action recognition as a function of speed is intuitively visualized in the curve plot of Fig.~\ref{Fig:conception} (b). The analysis results indicate that as the action velocity increases, the performance of ST-GCN, 2S-AGCN and ST-ST significantly deteriorates. This phenomenon can be attributed to the single-scale network structure used by these methods, which becomes particularly limiting when processing high-speed actions. However, when employing multi-scale networks, such as MS-G3D, HD-GCN, Infor-GCN, and CTR-GCN, this downward trend is effectively mitigated. Interestingly, our proposed method does not exhibit a significant impact on its action recognition performance due to changes in action velocity. On the contrary, when the action velocity increases, our method demonstrates a more pronounced advantage over other methods in terms of performance. This finding underscores the urgent need for insensitive features to velocity in action recognition tasks and highlights the potential application value of our method in this domain.

\subsection{Qualitative Results}
In the visual analysis, we elucidated the strategy of our VA-Transformer in selecting window sizes. As illustrated in Fig.~\ref{fig:visualization}, the first row presents frames extracted from action video clips. The second row showcases corresponding velocity information, reflecting the dynamic changes in behavior. The third row provides a detailed display of the weight allocation by MoWA for experts with different window sizes, revealing the model’s attentional focus on temporal information at varying speeds. Observations indicate that for daily behaviors, VA-Transformer tends to select the expert with larger window sizes (red bars) to capture a wealth of temporal information. Conversely, in athletic sports behaviors, the model favors smaller window sizes (cyan bars), focusing more on local temporal information. Even within a single action sample, as the speed varies, the model’s preference for window size also changes, with larger window sizes being selected at a smaller acceleration. This dynamic adjustment of window sizes reflects the model’s adaptability in handling behaviors of different speeds and is the fundamental reason for our method’s insensitivity to velocity.

\subsection{Ablation Study}
To achieve good performance of the Transformer architecture in skeleton-based action recognition tasks and ensure its good adaptability to recognizing actions at different speeds, we introduced a combined window attention mechanism of LWA (Local Window Attention) and SWA (Shifted Window Attention), as well as innovative designs such as a multi-scale window attention strategy and a mixture of expert module. To comprehensively evaluate the impact and effectiveness of these strategies, we conducted an ablation study. The results in Tab.~\ref{tab:ablation} clearly show that: i) the combined window attention mechanism of LWA and SWA significantly improves the Transformer’s capability to model temporal information by a $18.66\%$ increase; ii) the multi-scale window strategy further enhances the window attention mechanism’s ability to improve the model’s capability by a $2.74\%$ increase; iii) our MoWA can adaptively select the appropriate window scale, further enhancing the flexibility of the multi-scale window strategy by a $3.23\%$ increase.

\begin{table}[!t]
  \renewcommand\arraystretch{0.6}
  \tabcolsep=0.1cm
  \centering
  \scriptsize
    \begin{tabular}{p{1.4cm}cccc|c}
    \toprule
    Method & LWA & SWA & Multi-Scale & MoE & Accuracy(\%) \\
    \midrule
    GCN+Trans. &    &     &             &      & 58.4 \\
    GCN+Trans. &$\checkmark$&$\checkmark$&     &    & 69.3 \\
    GCN+Trans. &$\checkmark$&$\checkmark$&$\checkmark$&    &  71.2 \\
    \rowcolor[rgb]{ .906,  .902,  .902}GCN+Trans. &$\checkmark$&$\checkmark$&$\checkmark$&$\checkmark$& 73.5 \\
    \bottomrule
    \end{tabular}%
    \caption{Ablation study of each component within our method, conducted on the P2A dataset. The training exclusively utilizes the joint modality.}
    \label{tab:ablation}
\end{table}%

\paragraph{Impact of window size and number}
We employed an empirical approach, starting with windows of size 4 and incrementing the window size by a factor of 2. The performance of VA-Transformer under different numbers and sizes of windows is presented in Tab.~\ref{tab:window settings}. Through the results, we found that larger window sizes do not necessarily lead to better performance. In our experiments, when the window size is 8, VA-Transformer’s performance reaches a relatively ideal balance point. Moreover, the combination of multi-scale windows is not necessarily better with more scales. Excessive combinations of window sizes not only increase computational load but may also lead to model performance saturation.
\paragraph{How many experts per window?}
We explored the impact of the number of experts per window on the performance of VA-Transformer. As shown in Tab.~\ref{tab:window settings}, increasing the number of experts per window, regardless of whether the setting is single window or multi-scale window, can significantly enhance the model’s performance. However, considering the balance between computational resources and performance, we chose a configuration in the experiments where one expert corresponds to one window.

\begin{table}[t!]
  \renewcommand\arraystretch{0.6}
  \centering
  \scriptsize
    \begin{tabular}{ccccc}
    \toprule
    $\sharp$ Windows & Window Size &  $\sharp$ Experts/Window   & P2A & \makecell{Olympic\\Badminton}\\
    \midrule
   \multirow{6}{*}{single} & 4 & 1   & 68.1 & 67.5 \\ 
   & 4 & 2   & 70.3 & 69.8 \\
   & 8 & 1     & 69.3 & 69.0 \\ 
   & 8 & 2  & 71.5 & 70.6 \\ 
   & 16 & 1  & 69.1 & 69.0 \\
   & 16 & 2  & 71.5 & 70.7 \\ 
   \hdashline
  \multirow{4}{*}{multi} & \{4, 8\} & 1   & 71.9 & 71.7 \\ 
   & \{4, 8\} & 2   & 72.0 & 71.8 \\ 
   & \cellcolor[rgb]{ .906,  .902,  .902}\{4, 8, 16\}& \cellcolor[rgb]{ .906,  .902,  .902}1 & \cellcolor[rgb]{ .906,  .902,  .902}73.5  & \cellcolor[rgb]{ .906,  .902,  .902}73.2\\
    & \{4, 8, 16, 32\}& 1 & 73.3 & 73.1\\
    \bottomrule
    \end{tabular}%
  \caption{Impact of window size, number of windows, and number of experts per window on performance, evaluated on the P2A and Olympic Badminton datasets. The training exclusively utilizes the joint modality.}
  \label{tab:window settings}%
\end{table}%

\section{Conclusion}

In this paper, we investigate for the first time the impact of velocity on action recognition. Our pilot study reveals that current skeleton-based methods fail to maintain robust performance under scenarios of velocity changes.
To address this, we introduce the Velocity-Aware Action Recognition framework, which adapts the window size of attention dynamically by Mixture of Window Attention, employing larger windows for daily actions to capture extended temporal contexts, and smaller windows for rapid actions to focus on localized temporal variations. We hope our work inspires further research in developing the robustness and practicality of action recognition.

\newpage
\section{Acknowledgements}
This work is partially supported by the National Natural Science Foundation of China under the Grant No. 62476065.

\bibliography{aaai25}
\end{document}